\documentclass[letterpaper, 10 pt, conference]{IEEEtran}
\usepackage{times}

\usepackage{geometry}

\newgeometry{top=72pt, left=54pt, right=54pt, bottom=54pt}

\usepackage[numbers]{natbib}
\usepackage{multicol}
\usepackage[bookmarks=true]{hyperref}

\usepackage{multirow}
\usepackage{booktabs}
\usepackage{tabularx}

\usepackage{subcaption}
\usepackage{amsfonts,amssymb,amsmath,bm}
\usepackage{amsthm}
\usepackage{enumerate}
\usepackage{cite}
\usepackage[normalem]{ulem}
\usepackage{caption}
\usepackage{adjustbox}
\usepackage{booktabs}
\usepackage{subcaption}
\usepackage{algorithm}
\usepackage{algorithmic}

\makeatletter
\newcommand\notsotiny{\@setfontsize\notsotiny\@vipt\@viipt}
\makeatother
\usepackage{xcolor}

\usepackage{graphics} 
\usepackage{algorithm}
\usepackage{algorithmic}
\usepackage{dblfloatfix}

\DeclareMathOperator*{\argmax}{arg\,max}
\newtheorem{theorem}{Theorem}
\begin{document}

\title{Beyond Shortsighted Navigation: Merging Best View Trajectory Planning with Robot Navigation}


\author{\authorblockN{Srinath Tankasala}
\authorblockA{
stankasa@utexas.edu}
\and
\authorblockN{Roberto Mart\'in-Mart\'in}
\authorblockA{robertomm@cs.utexas.edu}
\and
\authorblockN{Mitch Pryor}
\authorblockA{mpryor@utexas.edu}
}


%

\maketitle

\begin{abstract}
Gathering visual information effectively to monitor known environments is a key challenge in robotics. To be as efficient as human surveyors, robotic systems must continuously collect observational data required to complete their survey task. Inspection personnel instinctively know to look at relevant equipment that happens to be ``along the way.''  In this paper, we introduce a novel framework for continuous long-horizon viewpoint planning, for ground robots, applied to tasks involving patrolling, monitoring or visual data gathering in known environments. Our approach to Long Horizon Viewpoint Planning (LHVP), enables the robot to autonomously navigate and collect environmental data optimizing for coverage over the horizon of the patrol. Leveraging a quadruped's mobility and sensory capabilities, our LHVP framework plans patrol paths that account for coupling the viewpoint planner for the arm camera with the mobile base's navigation planner. The viewpath optimization algorithm seeks a balance between comprehensive environmental coverage and dynamically feasible movements, thus ensuring prolonged and effective operation in scenarios including monitoring, security surveillance, and disaster response. We validate our approach through simulations and in the real world and show that our LHVP significantly outperforms naive patrolling methods in terms of area coverage generating information-gathering trajectories for the robot arm. Our results indicate a promising direction for the deployment of mobile robots in long-term, autonomous surveying, and environmental data collection tasks, highlighting the potential of intelligent robotic systems in challenging real-world applications.
\end{abstract}

\IEEEpeerreviewmaketitle

\section{Introduction}
This paper introduces a method for long-horizon viewpoint planning (LHVP) for conducting surveys and patrols with ground robots, such as quadrupeds. Robots have been increasingly integrated into essential tasks such as continuous monitoring of infrastructure, or evaluating for damages, leaks or corrosion. Notable applications include 3D modeling of industrial facilities, \citet{burri2012aerial}, and monitoring such as the Boston Dynamics Orbit system \citep{BostonDynamicsOrbit}. In many of these situations, the structures under examination are often represented through 3D models usually as high-resolution meshes or voxelized formats. Subsequently, robots are then redeployed to either enhance the resolution of these models to a higher accuracy or to conduct inspections for any alterations, potential dangers, or defects, such as cracks and corrosion. These applications involve surveying and monitoring tasks where the robot follows predetermined patrol paths, which can be set either by a human operator or an autonomous ground navigation planner. A key feature of such systems is the use of a camera mounted on a 6 Degrees of Freedom (DoF) arm (``eye-in-hand'' configuration) or a high resolution pan-tilt-zoom camera, which can be used for gathering image data for scene reconstruction. In this work we explore a planning system that synchronizes the robot's arm camera with its base movement for maximum scene coverage, eliminating unnecessary stops of the robot base to capture images. 

\begin{figure}[t]
    \begin{center}
        \includegraphics[width=0.45\textwidth]{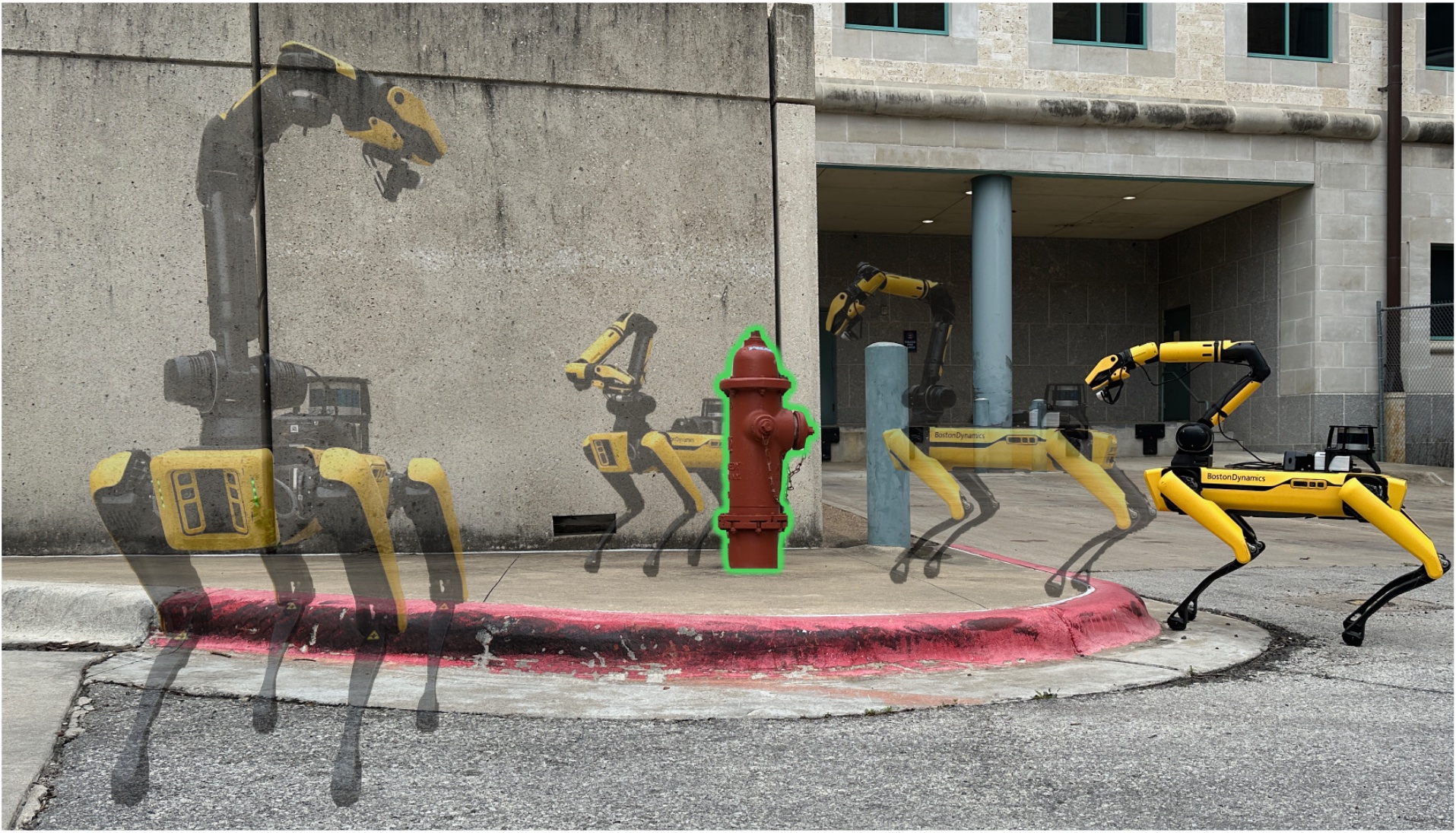}
        \captionof{figure}{Real-world execution of our long horizon view trajectory planner. The arm mounted camera moves in conditioned on the base navigation plan to effectively survey the environment without stopping.}
        \label{fig:spot_patrol}
    \end{center}
\end{figure}

Active perception, as defined by \citet{bajcsy2018revisiting}, is the ability of an embodied AI agent to understand the reasons for its sensing needs, choose its perception targets, and decide the manner, timing, and location of achieving this perception. In mobile robotics, various active perception techniques have been developed, particularly for motion creation through a Next-Best-View (NBV) strategy, as in \citep{Isler16, bircher2016receding, Naazare22}. Each NBV is selected using feedback from the current partial reconstruction of the environment. Most NBV techniques prioritize viewpoints based on information gain (IG) \citep{Isler16}, aiming to reduce uncertainty by exploring areas not yet observed. An in-depth comparison of different IG based approaches for NBV is available in \citet{Delmerico18}.

If the environment is known beforehand, then view planning is pre-computed and becomes an optimization problem with a cardinality constraint of choosing the best set of $N$ viewpoints \textit{VP}, i.e. $|VP| \leq N$, that maximizes coverage. It is considered as a non-negative monotone submodular function maximization which is NP-hard. The approximate solution for the viewpoint set, $VP$, is typically generated using the greedy algorithm as it has theoretical guarantees \citet{nemhauser1978analysis} and can usually be solved with $O(Nlog(N))$ complexity, example recursive greedy algorithm \citet{krause2014submodular, chekuri2005recursive, hepp2018plan3d}. However, applications of these greedy algorithms typically assume the agent can find a connecting tour between all viewpoints in $VP$, example a Traveling Salesman tour, such as \citet{bircher2016three}. This is applicable to drones which is a popular choice for surveying and photogrammetry applications \citep{srinath2022minimum, tankasala2022icuas}, and it is easier to generate feasible continuous trajectories for surveying in the Cartesian space for UAVs \citep{tankasala2022smooth, tankasala2022ecc, tankasala2023accelerating}. However, this is not applicable to ground robots which can have constraints between the base and the arm planners such as the Spot robot with an arm-mounted camera.

This work considers view planning in a known environment where the robot base travels along an input path that is pre-determined, possibly to complete a patrol or a primary task. We propose an efficient approach for visually informative motion generation that is conditioned on the robot base trajectory. In particular, we tackle the problem of coverage maximization of known scenes; this is a crucial problem to resolve for efficient surveying and monitoring using mobile robot agents. We demonstrate that by sequentially sampling view candidates for the arm mounted camera along the base patrol path, we account for the view path's feasibility and hence complete the surveying task as efficiently as possible. Since our planning system synchronizes the camera movement on the robot's arm with the movement of the robot base itself when optimizing for scene coverage, we ensure that the robot does not have to halt for the arm to reach the desired camera poses. There are many applications where the base may not be able to stop. Even if the robot does not achieve the maximum possible coverage during the first patrol, our planning approach can be used to plan camera view trajectories over multiple patrol cycles. We demonstrate a greedy search-based camera view path sampling and show that they work better than a viewpoint planner that do not consider the reachability constraints of the robot arm. 

The contributions in this work include 1) a novel perception motion generation approach in known environments that maximizes for scene coverage conditioned on robot base trajectories; 2) demonstrate the ability to balance information gain maximization over a long horizon while exploring executable arm motions; 3) and transfer the generated trajectories to real-world robots and generalize well over a wide range of objects and environment layouts.

\section{Camera motion generation for maximum scene coverage}

In our approach, we investigate scenarios where a mobile robot is deployed in a predefined environment, tasked with the objective of conducting a survey of a facility by performing visual data collection pertaining to specific objects integrated within the facility's layout. This is facilitated by the robot's array of components: a mobile base that ensures agile navigation across diverse scenes, an RGB-D camera for scene perception, and a 6DoF arm mounted on the base. This ``eye-in-hand'' setup is instrumental in capturing detailed visual data from a multitude of angles and perspectives and is commonly used for robotic surveying \citet{vasquez2014view}. The main difference from \citet{vasquez2014view} is that we only have control of the hand and not the robot base. The state of the robot base is defined by the position of its mobile base ($P^{base} \in SE(2)$), representing its planar movement. The camera's orientation and position ($P^{cam}$) are given by the end effector ($P^{eef} \in SE(3)$). This decoupling of the arm from the base is necessary, as the robot base's operational dynamics and planned survey paths are determined using navigation planners, such as \citet{BostonDynamicsOrbit}, and may be dictated by a different specified objective. 

Given the locations of the target objects of interest, indicated by $(L^1, L^2, L^3,...)$, a voxelized 3D occupancy map is used to plan safe movements and assess the information gain (IG) from candidate camera poses $P^{cam}$, \citet{Breyer22} for example. The primary goal is to enable the robot to efficiently collect data while accounting for the movement and velocity constraints on the arm. The motion generation for the robot arm, follows several guiding principles:

\begin{enumerate}
    \item IG is gathered over the entire horizon in which the target objects are visible from the robot's base locations.
    \item The base trajectory is known and discretized to calculate arm configurations for the camera poses
    \item Feasible camera view poses are sequentially generated in Cartesian space to cover all viewing orientations reachable by the camera as the robot traverses the base trajectory.
    
\end{enumerate}

The viewpoint optimization problem can be formulated as:
\begin{equation}
    \begin{split}
        &\max_{VP\subset Z} IG(VP)\\
        \text{s.t.}\ &cardinality(VP) \leq N
    \end{split}    
    \label{eq:high_level_obj}
\end{equation}
Where, $Z$ is the set of all camera pose trajectories of cardinality $\leq N$ that are feasible for the given path, $P$, of the robot base. This section introduces a comprehensive motion generation pipeline that adheres to these principles, involving the sampling of multiple viable camera paths for the robot to accomplish the surveying task.

\subsection{View candidates generation}
Given the location of the objects of interest in patrol rounds, we discretize the base path uniformly, for convenience, as shown in figure \ref{fig:algo_illustration} and determine suitable view candidates at each of those base locations. 

\begin{figure}[ht]
    \centering
    \includegraphics[width=0.45\textwidth]{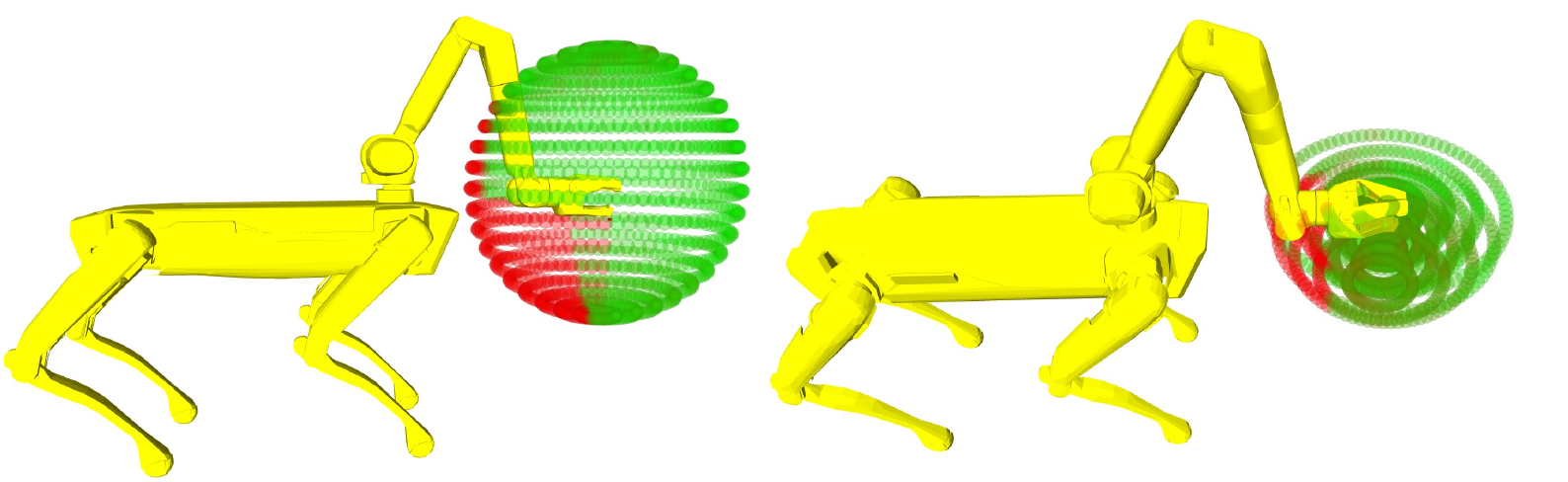}
    \caption{Reachability map of robot arm end effector in a single time step. The entire reachable surface in one time step is depicted on the left and cross-section of it is shown on the right. Pose dimensions reduced from 6D to 3D for visualization purposes.}
    \label{fig:pose_sampling}
\end{figure}

For the current base location $P_i$ and end effector pose $\tau_{i}$, the set \( V = \{\tau_{i+1}^1, \tau_{i+1}^2, \ldots, \tau_{i+1}^M\} \) represents the collection of potential camera poses for viewpoint selection at the next base location. Each view in this set is represented as \( \tau_{i+1}^j, j \in [1,\ldots, M] \), where \( \tau_{i+1}^j \in \mathbb{R}^3 \times \text{SO}(3) \) denotes the 6 dimensional sensor pose. To efficiently assess the potential candidates, the process utilizes a series of filters. Candidates that fail to meet the criteria of a filter are removed from the set of possible views. Figure \ref{fig:pose_sampling} illustrates the application of these filters. Given a desired average end effector velocity, $v_{eef}$, we uniformly sample poses in cartesian space and orientations that can be reached before the base gets to the next base pose ($P_{i+1}$). Following this initial filter, the remaining candidates are evaluated based on the positioning factor, which checks for any collisions between the candidate's position and the surrounding environment. Any candidate view that does not meet certain criteria during a ``filtering'' process is eliminated from the pool of potential views. Finally we use reachability of poses based on the current arm configuration $\tau_i$ and joint distance. The joint distance is calculated by assuming a trapezoidal velocity profile (TVP) based on maximum joint velocities(\(\Bar{\omega}_{\text{max}}\)) and maximum joint accelerations(\(\Bar{\alpha}_{\text{max}}\)) and the time step, $T_{step}$, to the next base location \(P_{i+1}\). The TVP bound for joint angles is described as shown in Eq. \eqref{eq:TVP}.

\begin{equation}
    TVP \left( \Bar{\omega}_{\text{max}}, \Bar{\alpha}_{\text{max}}, t \right) = 
    \begin{cases}
        \frac{\Bar{\alpha}_{\text{max}}}{2}\ T_q^2 + \Bar{\omega}_{\text{max}}\left(t-T_q\right) , \text{if } t > T_q\\
        \frac{\Bar{\alpha}_{\text{max}}}{2}\ t^2,\ \text{otherwise}        
     \end{cases}
    \label{eq:TVP}
\end{equation}
where, $T_q = \frac{\Bar{\omega}_{\text{max}}}{\Bar{\alpha}_{\text{max}}}$
\begin{algorithm}
\caption{View Candidate Generation and Filtering}
\textbf{Input:} Volumetric map (S), current arm and camera configuration ($q_{\text{curr}}$, $\tau_{\text{curr}}$), maximum end effector velocity $v_{\text{eef}}$, maximum joint velocities $\Bar{\omega}_{\text{max}}$, maximum joint accelerations $\Bar{\alpha}_{\text{max}}$, and time step between current and next base location sample $T_{step}$.

\begin{algorithmic}[1] 
\STATE $q_{\text{next}}^{max} \leftarrow q_{\text{curr}} + \text{TVP}(\Bar{\omega}_{\text{max}}, \Bar{\alpha}_{\text{max}}, T_{step})$
\STATE $V'\leftarrow\{ \tau\in V\mid||\text{IK}(\tau)-q_{\text{curr}}||\leq\text{TVP}(\Bar{\omega}_{\text{max}},\Bar{\alpha}_{\text{max}},T_{step})\}$               
\FOR{ \( \tau \in V' \)} 
     \STATE keep \( \tau \) if collision free
\ENDFOR
\RETURN $V'$
\end{algorithmic}
\label{alg:pose_filtering}
\end{algorithm}

The reachability map in Fig \ref{fig:pose_sampling} illustrates the robot arm's end effector reachability within a single time step, using $T_{step}=1s$ at $v_{eef}=0.25m/s$ using algorithm \ref{alg:pose_filtering}. The left image shows the reachable surface (a sphere of radius $v_{eef}\times T_{step}$), while the right image presents a cross-section of it showing all the sampled poses. The 3D voxels are color-coded based on the number of reachable orientations at each location, with green voxels indicating a higher number and red voxels indicating a lower number of reachable orientations. This visualization helps to understand the robot arm's capability in terms of reachable positions and orientations within one time step.



\subsection{Information gain and view path computation}

The objective of a next-best-view planner is to select the viewpoint that maximizes the total number of surface voxels observed on the object(s) of interest. The best camera view path on the other hand maximizes the aggregate number of voxels observed along the set of views $VP^*$. The selection of the world representation, the positioning of the sensor, and the calculation of the gain of information are vital elements for any active perception task. 

\begin{figure}[ht]
    \centering
    \begin{subfigure}[b]{0.45\textwidth}
        \includegraphics[width=\textwidth]{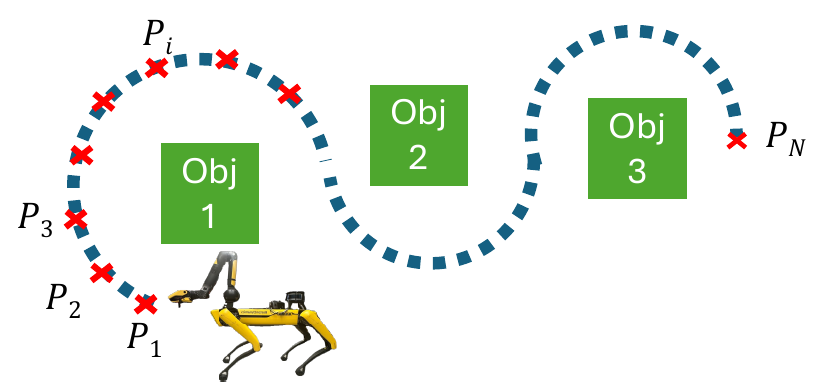}
    \end{subfigure}
    \hfill
    \begin{subfigure}[b]{0.49\textwidth}
        \includegraphics[width=\textwidth]{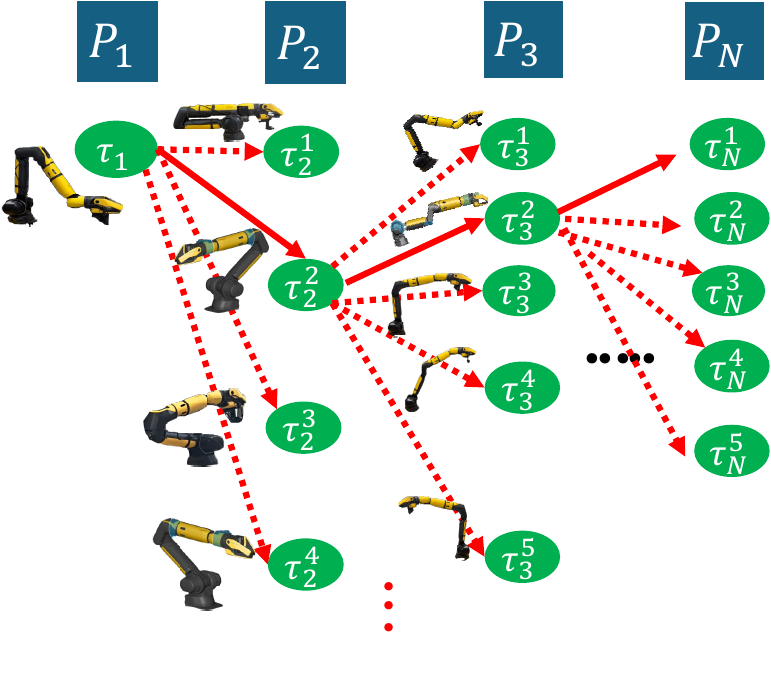}
    \end{subfigure}
    \caption{Greedy IG search along robot base path (dotted blue line) uniformly discretized as \(P_1, P_2,..., P_N\). Reachable camera pose candidates,\(\tau_{i+1}^1, \tau_{i+1}^2,..., \tau_{i+1}^M\),  are sequentially sampled from the current end effector pose, $\tau_i^j$. Greedy search traverses viewpoint graph along nodes of maximum marginal IG (solid red arrow path)}
    \label{fig:algo_illustration} 
\end{figure}

Volumetric models are widely used for visualizing 3D objects, as they offer a compact way of space encoding. We use a dense voxelized representation of the scene and compute the information gain to determine the utility of any given viewpoint using ray casting. The greedy viewpath sampling algorithm expands a tree of possible viewpoints sequentially. At each time step, the algorithm selects the candidate pose with the highest marginal Information Gain (IG) value. The IG value of a pose $\tau_i^j$ at time step $i$ and for pose $j$ is denoted by $IG(\tau_i^j)$. The process can be visualized as shown in figure \ref{fig:algo_illustration}.

The objective is to select a view ($\tau$) from the set ($V$) that maximizes the information gain (IG) over the traversed viewpoint set $VP$. An elementary method to approximately maximize monotone submodular functions under cardinality constraints as in Eq. \eqref{eq:high_level_obj}, involves the greedy algorithm \citep{krause2014submodular}. This process begins with an initial empty set $VP_0$, and in each iteration $i$, it incorporates the element that offers the maximum increase according to the discrete derivative $IG\left(e|VP^{i-1}\right)$(ties broken randomly):
\begin{equation}
    VP^{i} = VP^{i-1}\cup \left\{ \argmax_{e} IG( e|VP^{i-1} ) \right\}
    \label{eq:greedy_max}
\end{equation}

A notable result from \citet{nemhauser1978analysis} established that the greedy algorithm gives an effective approximation for the optimal solution of the NP-hard submodular function optimization problem of Eq. \ref{eq:high_level_obj}.

\begin{theorem}[\citet{nemhauser1978analysis}]
For a non-negative monotone submodular function \( f: 2^V \rightarrow \mathbb{R}_+ \), let \( \{VP^i\}_{i\geq 0} \) be the greedily selected sets defined in Eq. \eqref{eq:greedy_max}. Then for all positive integers \( k \) and \( \ell \), 
\[
f(VP^{\ell}) \geq (1 - e^{-\ell/k}) \max_{VP: |VP|\leq k} f(VP).
\]
In particular, for \( \ell = k \), 
\[
f(VP^k) \geq (1 - 1/e) \max_{VP: |VP|\leq k} f(VP).
\]
\end{theorem}

Using the above strategy, we employ the greedy search approach to determine the optimal view path. The flow of the process is illustrated in Figure \ref{fig:algo_illustration}. Initially, a collection of potential views is generated by uniformly sampling poses in the Cartesian space based on the current end effector position and orientation \(\tau_i\). Subsequently, these views undergo prioritization based on a utility function (marginal IG), with the optimal camera pose ($\tau^*$) being the one that yields the highest utility value ($IG\left(\tau^*|VP^{i-1}\right)$). This way, we consider not just the Next Best View but account for IG over the horizon of the robot's base trajectory. 

\begin{algorithm}
\caption{Long Horizon Camera View Planning Algorithm using Greedy sampling}
\begin{algorithmic}[1]
\STATE Initialize the set of view poses \( VP \) with the starting view point $\tau_1$ and base pose $P_1$
\WHILE{not at the maximum depth of tree}
    \STATE Determine feasible candidate poses, $V'$, at next base location $P_{i+1}$ using algorithm 1
    \STATE \( IG_{\text{max}}, \tau_{\text{best}} \gets -\infty, \text{null}  \)
    \FOR{\( \tau_i^j \in V' \)}
        \STATE Calculate \( IG(VP \cup \{\tau_i^j\}) \)
        \IF{\( IG(VP \cup \{\tau_i^j\}) > IG_{\text{max}} \)}
           \STATE \( IG_{\text{max}} \gets IG(VP \cup \{\tau_i^j\}) \)
            \STATE \( \tau_{\text{best}} \gets \tau_i^j \)
        \ENDIF
    \ENDFOR
    \STATE Add \( \tau_{\text{best}} \) to \( VP \)
    \STATE Move to \( \tau_{\text{best}} \) and set as the current view point
\ENDWHILE
\STATE \textbf{return} The set of view poses \( VP \)
\label{alg:greedy_IG}
\end{algorithmic}
\end{algorithm}

\section{Experiments}
We test the effectiveness of our planner in both simulation and real world scenarios. The primary metric for evaluating the proposed planner is the coverage of the reconstruction within the region of interest. A view is considered informative if it sees any surfaces of the objects of interest. We assess this in our IG computation by performing a ray casting and downsampling the ray density by 10 to make it run fast.

\subsection{Simulation}
\begin{figure}[ht]
    \centering
    \includegraphics[width=0.45\textwidth]{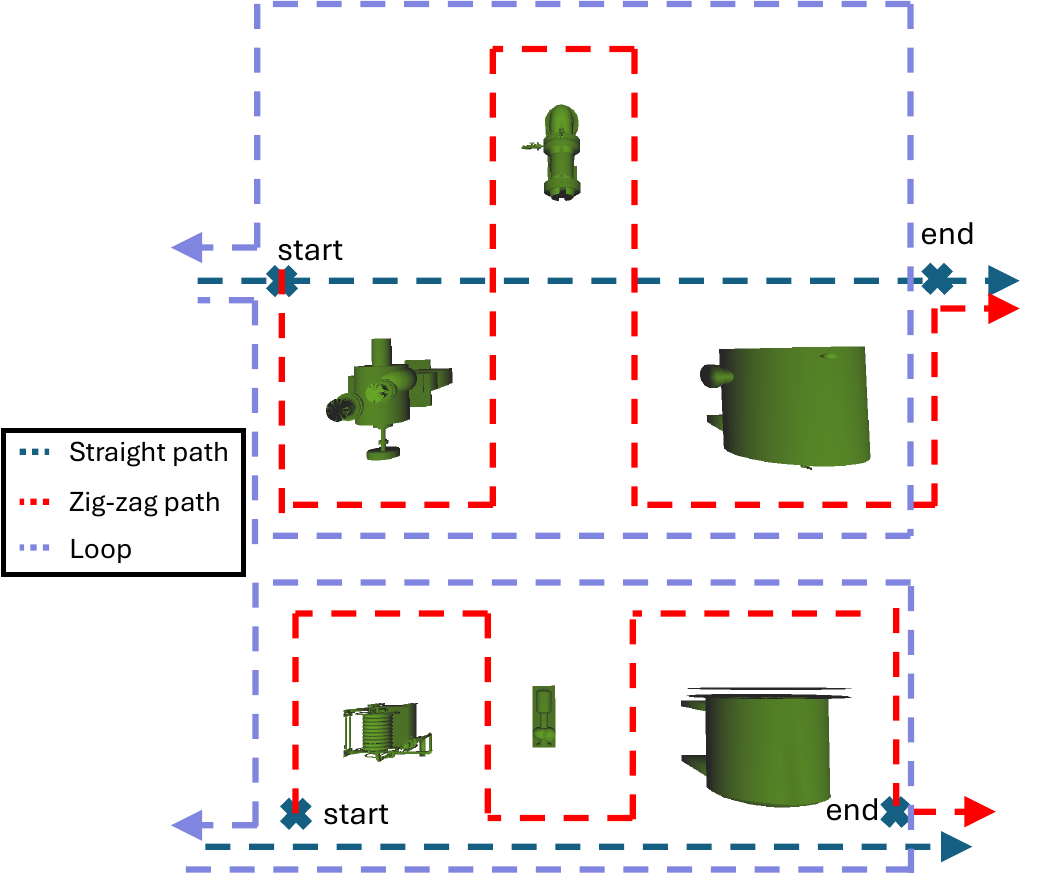}
    \caption{Different environment configurations, triangle layout (top) and linear layout(bottom), and base paths (Straight, Loop, Zig-zag) used to evaluate our long horizon view trajectory planner}
    \label{fig:layouts} 
\end{figure}

\begin{figure}
    \centering
    \includegraphics[width=0.45\textwidth]{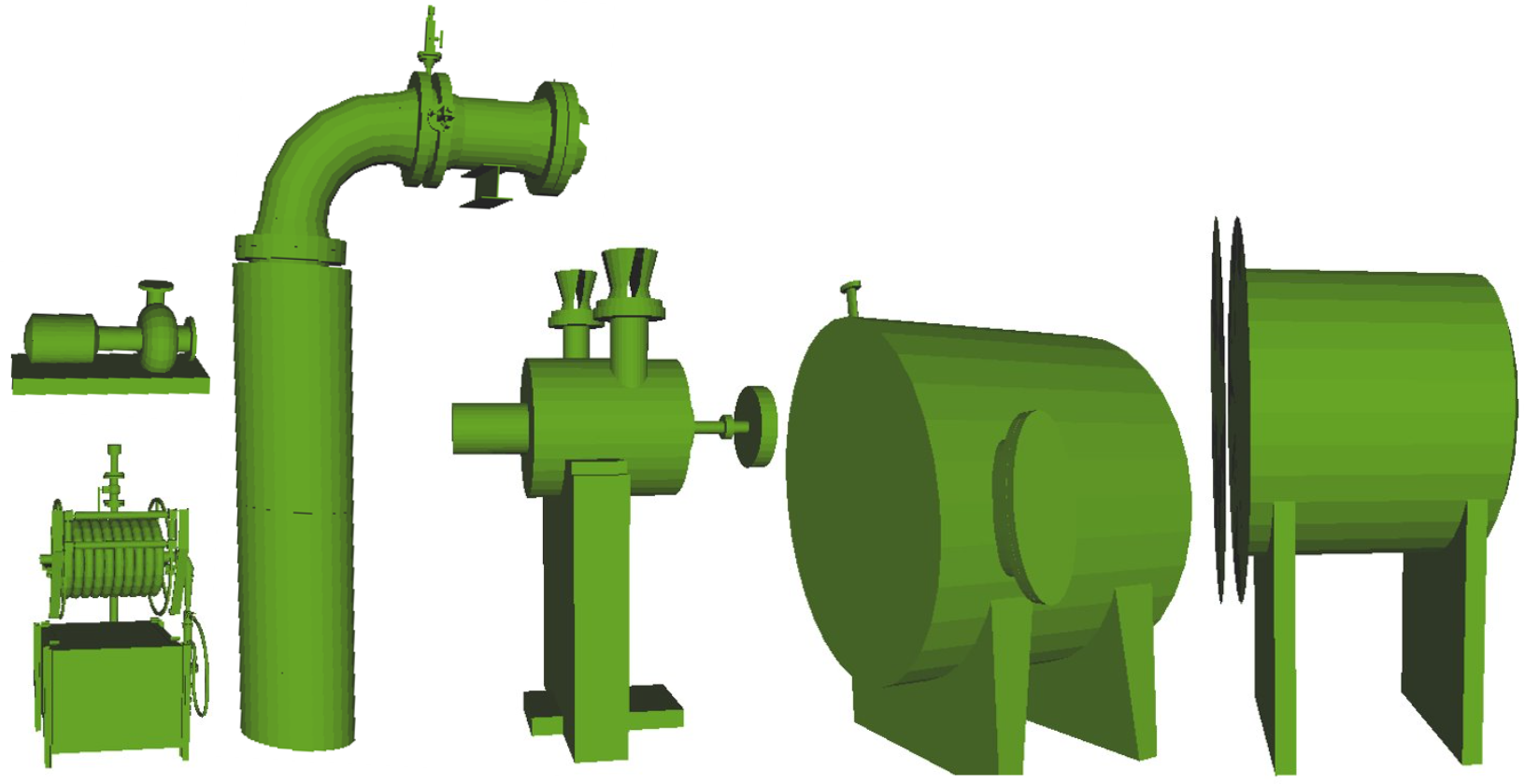}
    \caption{Sample assets used as survey targets of interest (to scale)}
    \label{fig:assets}
\end{figure}

Our simulation places multiple objects in different configurations, with different possible tours by the robot (as shown in Fig. \ref{fig:layouts} and \ref{fig:assets}). We create 2 configurations and 3 unique robot base paths around the objects. We utilize different assets to ensure that the algorithm is evaluated over multiple object shapes and measure how well the selected view paths offer varied angles on the objects. We use six models commonly seen in refinery settings, as seen in Fig. \ref{fig:assets}. Gazebo was used for the simulations and stereo processing is conducted in ROS. We use the marginal IG function (Eq. \eqref{eq:greedy_max}) as the utility function to perform the greedy search along the view path graph. 

To measure reconstruction quality by surface coverage, we evaluate the reconstructed point clouds against the original model's point cloud. For every point in the original model, we find the nearest point in the reconstructed model. If this point is within a specified registration distance, the surface point from the original model is deemed ``captured''. For simulations, we choose the registration distance to be equal to the unit size of the voxel (5 cm). Surface coverage ($C$) is calculated as the proportion of these captured surface points relative to the total surface points of the model:

\begin{equation}
    C = \frac{\text{observed surface points}}{\text{total points on original model surface}}
    \label{eq:coverage}
\end{equation}

We average the coverage value over different assets so the coverage estimate is only affected by the object layouts and the robot's base path. The maximum and minimum coverage values seen over individual objects are also reported (Table \ref{tab:complex_scene}) to understand the variance in performance.

We compare our method with a baseline heuristic approach that looks only at the nearest object. In this approach, Looking at the nearest object points the camera towards the object that is closest to the robot along the base path. If multiple objects nearby are equidistant from the base path, then the object center is arbitrarily chosen among them. A comparison of our method to the baseline is shown in Table \ref{tab:complex_scene}.

\begin{table}[htbp]
\centering
\caption{Coverage study}
\begin{tabular}{l*{4}{c}}
\toprule
Base path & \multicolumn{2}{c}{Triangle layout} & \multicolumn{2}{c}{Linear layout} \\
\cmidrule(r){2-3} \cmidrule(l){4-5}
LHVP (ours) & Mean & (Max, Min) & Mean & (Max, Min) \\
\midrule
Zig-Zag & 78.2\% & $\left(91.5\%,59.3\%\right)$ & 62.9\% & $\left(77.5\%,30.7\%\right)$ \\
Straight & 40.1\% & $\left(53.1\%,9.6\%\right)$ & 49.5\% & $\left(55.1\%,31.4\%\right)$ \\
Loop & 59.7\% & $\left(81.3\%,33.2\%\right)$ & 61.3\% & $\left(74.7\%,51.9\%\right)$ \\
\midrule
See Nearest &&&&  \\
(baseline) &&&&\\
\midrule
Zig-Zag & 64.5\% & $\left(70.3\%,56.1\%\right)$ & 51.8\% & $\left(64.4\%,25.9\%\right)$ \\
Straight & 31.8\% & $\left(41.2\%,15.5\%\right)$ & 28.7\% & $\left(36.1\%,19.3\%\right)$ \\
Loop & 39.6\% & $\left(56.2\%,26.4\%\right)$ & 45.9\% & $\left(70.2\%,30.3\%\right)$ \\
\bottomrule
\end{tabular}
\label{tab:complex_scene}
\end{table}
We run the planner with $v_{base} = 0.5m/s$, $T_{step}=2s$ and $v_{eef}=0.4$ in all simulation results. Focusing on the mean coverage values in Table \ref{tab:complex_scene}, we clearly see that our long horizon viewpath planner achieves greater coverage regardless of the chosen layout and base path followed by the robot. This is especially true in the Zig-Zag base path where it is important for the robot to keep switching between viewing different objects to collect as much information as possible. Looking only at the nearest object caused that method to completely miss faces of certain objects. Our planning approach is also able to focus enough attention across objects leading to a higher $(Max,Min)$ coverage individually, compared to focusing only on the nearest object.

\subsection{Ablation studies}
Table \ref{tab:algorithm1} presents the ablation of not considering joint reachability constraints during sequential viewpoint sampling, i.e. not using Algorithm \ref{alg:pose_filtering}. Without filtering unreachable poses in the sequential viewpoint sampling process, our greedy search traverses a viewpath along the graph that is not executable on the real robot, leading to significantly poorer performance (78\% vs 47\% in zig-zag case for example). This highlights the importance of taking the arm velocity constraints into account when planning the viewpath.
\begin{table}[htbp]
\centering
\caption{Ablation - joint reachability}
\begin{tabular}{l*{3}{c}}
\toprule
View sampling  & Triangle layout  & Linear layout  \\
method & & \\
\midrule
Greedy search w/out joint \\ distance filter & 47.45\%& 36.32\% \\
Greedy search w/ joint\\ distance filter & 78.25\% & 62.85\% \\
\bottomrule
\end{tabular}
\label{tab:algorithm1}
\end{table}

\begin{figure}
    \centering
    \includegraphics[width=0.5\textwidth]{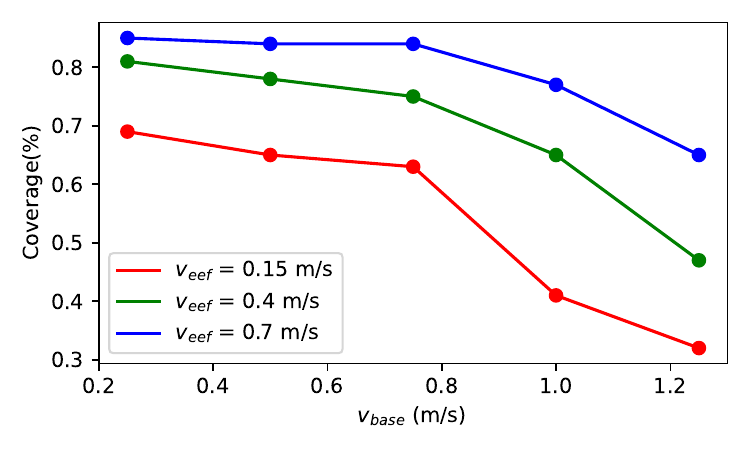}
    \caption{Effect of base and arm velocity on coverage of generated paths}
    \label{fig:velocity_effect}
\end{figure}

\begin{figure}
    \centering
    \includegraphics[width=0.5\textwidth]{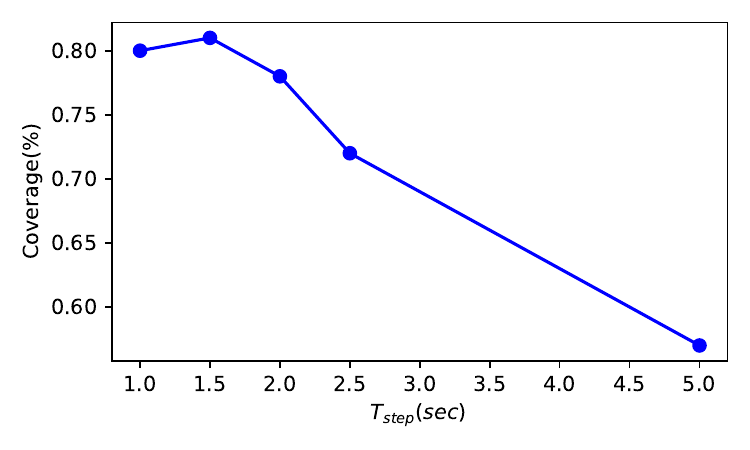}
    \caption{Effect of reducing time step between base positions}
    \label{fig:time_step_effect}
\end{figure}

We study the effect of the main hyperparameters used to generate the viewpath graph, namely $v_{eef}, v_{base}$ and $T_{step}$. These parameters directly affect the viewpoint graph that is traversed greedily (Fig \ref{fig:pose_sampling} and \ref{fig:algo_illustration}). Fig \ref{fig:velocity_effect} depicts the  coverage possible (at $T_{step}=2s$) as a function of $v_{base}$ along a zig-zag path in a linear enivronment layout. We observe that as the base moves slower, it is able to achieve higher coverage of the environment which is expected. The same effect occurs when $T_{step}$ is reduced as shown in Fig \ref{fig:time_step_effect}. Reducing $v_{base}$ or $T_{step}$ has the effect of increasing the length of the viewpath graph (Fig \ref{fig:algo_illustration} leading to more views collected for a given survey. 
If however, we reduce the average end effector velocity $v_{eef}$ (motion blur purposes), then the planned viewpath has lower coverage. This arises from the fact that the sequential sampling (Fig \ref{fig:pose_sampling}) gets constrained to a smaller volume leading to less diverse view candidates considered at each time step.

We further study the difference between the maximum coverage if the robot base could stop at each location $P_1, P_2,...,P_N$ along the base path. Halting the robot base increases the amount of reachable poses at each base location $P_i$ this consequently leads to better overall coverage but increases survey time. Table \ref{tab:coverage_rate} shows the mean coverage rate (\%/s) and total time for each survey. While the total coverage is higher when the base halts along the path, the rate at which the environment is covered is higher when the robot doesn't halt along the base path. This would make our approach useful in cases where the battery time is limited or in time-critical missions.
\begin{table}[ht]
\centering
\caption{Coverage rate with and without stopping the robot base}
\begin{tabular}{lcc}
\hline
\textbf{Base Path} & w/out stopping & w/ stopping \\
\hline
Zig-Zag & (1.49\%/s, 42s) & (1.34\%/s,52s) \\
Straight & (2.72\%/s,18s) & (2.34\%/s,24s) \\
Loop & (1.27\%/s,48s) & (1.14\%/s,59s) \\
\hline
\end{tabular}
\label{tab:coverage_rate}
\end{table}

In our final ablation study, we compare the coverage if we only use the images taken at the viewpoints for reconstruction vs utilizing the entire camera stream. The results are shown in Table \ref{tab:all_images}. In certain cases, the difference seems small (59.7\% vs 47.3\% in the loop path) suggesting that the arm may be undergoing large jumps in joint angles, causing the camera to not be pointed at the objects of interest in the intermediate steps. Choosing a smaller $T_{step}$ helps ensure a smoother trajectory and can capture data in the intermediate frames to further improve the reconstruction quality of the scene.
\begin{table}[ht]
\centering
\caption{Ablation-using entire camera stream}
\begin{tabular}{lcc}
\hline
\textbf{Base Path} & use all images & use only viewpoint images  \\
\hline
Zig-Zag & 78.25\% & 69.8\% \\
Straight & 40.1\% & 23.41\% \\
Loop & 59.7\% & 47.3\% \\
\hline

\hline
\end{tabular}
\label{tab:all_images}
\end{table}

\subsection{Real world reconstruction}

\begin{figure}[t]  
\centering
\subcaptionbox{Real world environment test}[0.45\textwidth]{
  \includegraphics[width=0.45\textwidth]{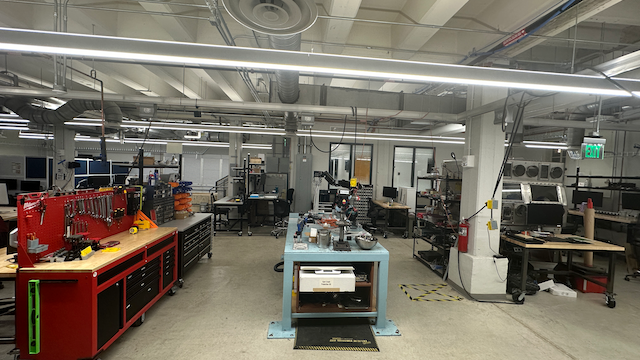}
}
\hfill
\subcaptionbox{Reconstruction comparison using our planner (top) vs ground truth (bottom)}[0.45\textwidth]{
  \includegraphics[width=0.45\textwidth]{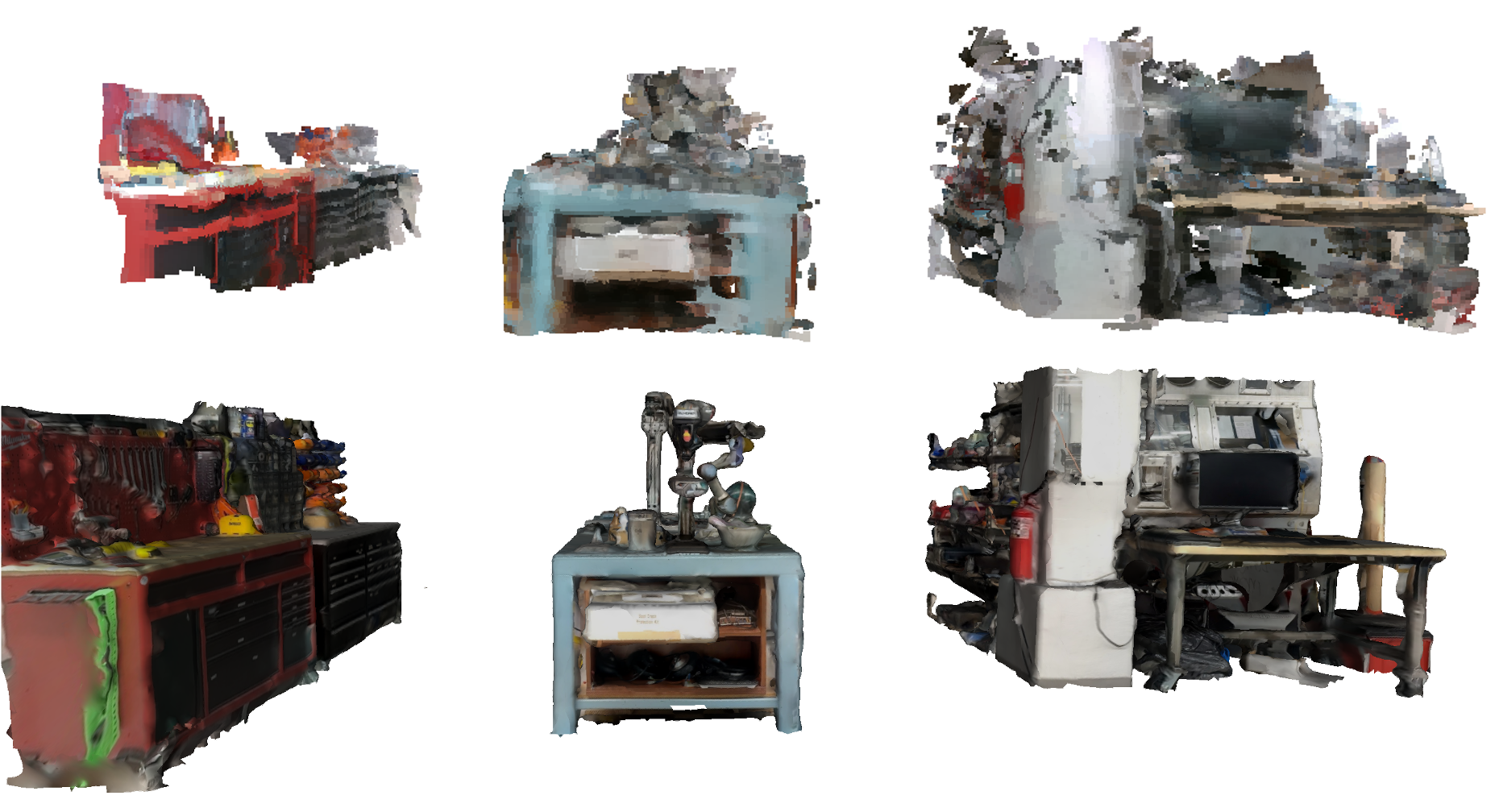}
}
\caption{Reconstruction quality comparison experiments on real robot, (a) Lab environment used for testing our planner on a Zig-zag path and (b) Point cloud generated from data collected using our planner compared to ground truth high quality scan}
\label{fig:AHG}
\end{figure}

In this experiment, we used the eye-in-hand configuration of the Spot robot to reconstruct the lab space shown in Figure~\ref{fig:AHG} to demonstrate the approach can efficiently handle real-world environments. We build a TSDF representation from the image stream taken by the robot arm as it executes the desired trajectory, and setting the registration distance threshold at 4 cm and the TSDF grid size to 4 cm. 

\begin{table}[ht]
\centering
\caption{Experimental Parameters}
\begin{tabular}{lcc}
\hline
TSDF size & $l$ & 10m \\
TSDF voxel grid resolution & R & 200 \\
Max View candidates per base pose & M & 980 \\
Base path resolution & N & 25 \\
Base velocity & $\nu_{base}$ & 0.35 m/s\\
Max. arm velocity & $\nu_{eef}$ & 0.65 m/s\\
\hline
\end{tabular}
\label{tab:experimental_params}
\end{table}

\begin{table}[ht]
\centering
\caption{Reconstruction coverage of real world environment using our camera view path planner}
\begin{tabular}{lcc}
\hline
\textbf{Base Path} & $T_{step}= 1\ sec$ & $T_{step}= 2.0\ sec$ \\
\hline
Method - LHVP (ours)\\
\hline
Zig-Zag & 57.8\% & 50.6\% \\
Straight & 20.48\% & 19.32\% \\
Loop & 47.4\% & 48.5\% \\
\hline
Method - See nearest\\
\hline
Zig-Zag & 32.4\% & 29.9\% \\
Straight & 16.7\% & 18.2\% \\
Loop & 30.2\% & 31.4\% \\
\hline
\end{tabular}
\label{tab:moving_base}
\end{table}

The reconstructed map from the data collected using our planner is shown in Fig \ref{fig:AHG} and compared to a high-fidelity point cloud. The reason for the noisy reconstruction (especially of the objects on the table) is due to motion blurring as the robot walks. The camera experiences vibrations due to Spot's walking gait leading to blurry captures despite tuning the hyperparameters as shown in Table \ref{tab:experimental_params}. Our method still performs better than the baseline (see nearest) which has the same reconstruction issue. This reconstruction issue is unrelated to the proposed planning principle in this work. Thus, our experimental results are still significant and demonstrate the effectiveness of our approach. Optimizing the robot gait for smoother image data acquisition could be an avenue for future work.

\section{Conclusion} 
\label{sec:conclusion}
This work presented a new view planning approach for long horizon camera trajectory generation that synchronizes the sensor trajectory to that of the robot base. This approach is derived from the greedy search method commonly used in viewpath optimization and is adapted to this problem by our novel view sampling approach. Our approach generates a graph, along the base path, where executable view candidates are sequentially sampled and the graph is traversed based on maximum marginal information gain. This results in a camera viewpath that maximizes IG over a long horizon while efficiently searching over the set of all possible view paths. We successfully demonstrate the practicality and effectiveness of our approach on a Spot robot in both simulation and the real world. Our ablation studies highlight how different hyperparameter choices in the planner affect the coverage performance. The developed method performs better than the heuristic baseline of looking at the nearest object across multiple environment settings. Moving forward, we intend to delve into more refined search optimization methodologies, such as recursive greedy search, to develop camera trajectories in scenarios where the view planner lacks control over the robot base trajectory.


\bibliographystyle{plainnat}
\bibliography{references}

\begin{thebibliography}{19}
\providecommand{\natexlab}[1]{#1}
\providecommand{\url}[1]{\texttt{#1}}
\expandafter\ifx\csname urlstyle\endcsname\relax
  \providecommand{\doi}[1]{doi: #1}\else
  \providecommand{\doi}{doi: \begingroup \urlstyle{rm}\Url}\fi

\bibitem[Bajcsy et~al.(2018)Bajcsy, Aloimonos, and Tsotsos]{bajcsy2018revisiting}
Ruzena Bajcsy, Yiannis Aloimonos, and John~K Tsotsos.
\newblock Revisiting active perception.
\newblock \emph{Autonomous Robots}, 42:\penalty0 177--196, 2018.

\bibitem[Bircher et~al.(2016{\natexlab{a}})Bircher, Kamel, Alexis, Burri, Oettershagen, Omari, Mantel, and Siegwart]{bircher2016three}
Andreas Bircher, Mina Kamel, Kostas Alexis, Michael Burri, Philipp Oettershagen, Sammy Omari, Thomas Mantel, and Roland Siegwart.
\newblock Three-dimensional coverage path planning via viewpoint resampling and tour optimization for aerial robots.
\newblock \emph{Autonomous Robots}, 40:\penalty0 1059--1078, 2016{\natexlab{a}}.

\bibitem[Bircher et~al.(2016{\natexlab{b}})Bircher, Kamel, Alexis, Oleynikova, and Siegwart]{bircher2016receding}
Andreas Bircher, Mina Kamel, Kostas Alexis, Helen Oleynikova, and Roland Siegwart.
\newblock Receding horizon" next-best-view" planner for 3d exploration.
\newblock In \emph{2016 IEEE international conference on robotics and automation (ICRA)}, pages 1462--1468. IEEE, 2016{\natexlab{b}}.

\bibitem[Breyer et~al.(2022)Breyer, Ott, Siegwart, and Chung]{Breyer22}
Michel Breyer, Lionel Ott, Roland Siegwart, and Jen~Jen Chung.
\newblock Closed-loop next-best-view planning for target-driven grasping, 2022.
\newblock URL \url{arxiv.org/abs/2207.10543}.

\bibitem[Burri et~al.(2012)Burri, Nikolic, H{\"u}rzeler, Caprari, and Siegwart]{burri2012aerial}
Michael Burri, Janosch Nikolic, Christoph H{\"u}rzeler, Gilles Caprari, and Roland Siegwart.
\newblock Aerial service robots for visual inspection of thermal power plant boiler systems.
\newblock In \emph{2012 2nd int. conf. on applied robotics for the power industry (CARPI)}, pages 70--75. IEEE, 2012.

\bibitem[Chekuri and Pal(2005)]{chekuri2005recursive}
Chandra Chekuri and Martin Pal.
\newblock A recursive greedy algorithm for walks in directed graphs.
\newblock In \emph{46th annual IEEE symposium on foundations of computer science (FOCS'05)}, pages 245--253. IEEE, 2005.

\bibitem[Delmerico et~al.(2018)Delmerico, Isler, Sabzevari, and Scaramuzza]{Delmerico18}
Jeffrey Delmerico, Stefan Isler, Reza Sabzevari, and Davide Scaramuzza.
\newblock A comparison of volumetric information gain metrics for active 3d object reconstruction.
\newblock \emph{Autonomous Robots}, 42\penalty0 (2), 2018.

\bibitem[Dynamics(2024)]{BostonDynamicsOrbit}
Boston Dynamics.
\newblock Robot fleet management software from boston dynamics gives you a single view of your site, robots, and equipment.
\newblock https://bostondynamics.com/products/orbit/, 2024.

\bibitem[Hepp et~al.(2018)Hepp, Nie{\ss}ner, and Hilliges]{hepp2018plan3d}
Benjamin Hepp, Matthias Nie{\ss}ner, and Otmar Hilliges.
\newblock Plan3d: Viewpoint and trajectory optimization for aerial multi-view stereo reconstruction.
\newblock \emph{ACM Transactions on Graphics (TOG)}, 38\penalty0 (1):\penalty0 1--17, 2018.

\bibitem[Isler et~al.(2016)Isler, Sabzevari, Delmerico, and Scaramuzza]{Isler16}
Stefan Isler, Reza Sabzevari, Jeffrey~A. Delmerico, and Davide Scaramuzza.
\newblock An information gain formulation for active volumetric 3d reconstruction.
\newblock \emph{2016 IEEE International Conference on Robotics and Automation (ICRA)}, pages 3477--3484, 2016.

\bibitem[Krause and Golovin(2014)]{krause2014submodular}
Andreas Krause and Daniel Golovin.
\newblock Submodular function maximization.
\newblock \emph{Tractability}, 3\penalty0 (71-104):\penalty0 3, 2014.

\bibitem[Naazare et~al.(2022)Naazare, Rosas, and Schulz]{Naazare22}
Menaka Naazare, Francisco~Garcia Rosas, and Dirk Schulz.
\newblock Online next-best-view planner for 3d-exploration and inspection with a mobile manipulator robot.
\newblock \emph{{IEEE} Robotics and Automation Letters}, 7\penalty0 (2):\penalty0 3779--3786, apr 2022.
\newblock \doi{10.1109/lra.2022.3146558}.

\bibitem[Nemhauser et~al.(1978)Nemhauser, Wolsey, and Fisher]{nemhauser1978analysis}
George~L Nemhauser, Laurence~A Wolsey, and Marshall~L Fisher.
\newblock An analysis of approximations for maximizing submodular set functions—i.
\newblock \emph{Mathematical programming}, 14:\penalty0 265--294, 1978.

\bibitem[Srinath et~al.(2022)Srinath, Can, and Mitch]{srinath2022minimum}
Tankasala Srinath, Pehlivanturk Can, and Pryor Mitch.
\newblock Minimum time trajectory generation for surveying using uavs.
\newblock \emph{arXiv preprint arXiv:2202.11297}, 2022.

\bibitem[Tankasala and Pryor(2023)]{tankasala2023accelerating}
Srinath Tankasala and Mitch Pryor.
\newblock Accelerating trajectory generation for quadrotors using transformers.
\newblock In \emph{Learning for Dynamics and Control Conference}, pages 600--611. PMLR, 2023.

\bibitem[Tankasala et~al.(2022{\natexlab{a}})Tankasala, Pehlivanturk, Bakolas, and Pryor]{tankasala2022ecc}
Srinath Tankasala, Can Pehlivanturk, Efstathios Bakolas, and Mitch Pryor.
\newblock Generating smooth near time-optimal trajectories for steering drones.
\newblock In \emph{2022 European Control Conference (ECC)}, pages 1484--1490, 2022{\natexlab{a}}.
\newblock \doi{10.23919/ECC55457.2022.9838451}.

\bibitem[Tankasala et~al.(2022{\natexlab{b}})Tankasala, Pehlivanturk, Bakolas, and Pryor]{tankasala2022smooth}
Srinath Tankasala, Can Pehlivanturk, Efstathios Bakolas, and Mitch Pryor.
\newblock Smooth time optimal trajectory generation for drones.
\newblock \emph{arXiv preprint arXiv:2202.09392}, 2022{\natexlab{b}}.

\bibitem[Tankasala et~al.(2022{\natexlab{c}})Tankasala, Pehlivanturk, and Pryor]{tankasala2022icuas}
Srinath Tankasala, Can Pehlivanturk, and Mitch Pryor.
\newblock Near minimum time trajectory planning for surveying using uavs.
\newblock In \emph{2022 International Conference on Unmanned Aircraft Systems (ICUAS)}, pages 873--880, 2022{\natexlab{c}}.
\newblock \doi{10.1109/ICUAS54217.2022.9836132}.

\bibitem[Vasquez-Gomez et~al.(2014)Vasquez-Gomez, Sucar, and Murrieta-Cid]{vasquez2014view}
J~Irving Vasquez-Gomez, L~Enrique Sucar, and Rafael Murrieta-Cid.
\newblock View planning for 3d object reconstruction with a mobile manipulator robot.
\newblock In \emph{2014 IEEE/RSJ International Conference on Intelligent Robots and Systems}, pages 4227--4233. IEEE, 2014.

\end{thebibliography}

\end{document}